\documentclass[runningheads]{llncs}
\usepackage[numbers]{natbib}
\usepackage{graphicx}
\usepackage{caption}
\usepackage{subcaption}
\usepackage{multirow}
\usepackage[utf8]{inputenc}
\usepackage{amsmath}
\usepackage{amssymb}
\usepackage{lipsum}
\usepackage{hyperref}

\begin{document}
\title{Clustering Object-Centric Event Logs}
%
%
\author{Anahita Farhang Ghahfarokhi\inst{1}\and
Fatemeh Akoochekian\inst{1} \and
Fareed Zandkarimi\inst{2}\and
Wil M.P. van der Aalst\inst{1}}
\authorrunning{Anahita Farhang Ghahfarokhi et al.}
%
\institute{Process and Data Science, RWTH Aachen University, Aachen, Germany \and
Chair of Enterprise Systems, University of Mannheim, Mannheim, Germany
\email{farhang@pads.rwth-aachen.de}}
\maketitle              
\begin{abstract}
Process mining provides various algorithms to analyze process executions based on event data. Process discovery, the most prominent category of process mining techniques, aims to discover process models from event logs, however, it leads to spaghetti models when working with real-life data. Therefore, several clustering techniques have been proposed on top of traditional event logs (i.e., event logs with a single case notion) to reduce the complexity of process models and discover homogeneous subsets of cases. Nevertheless, in real-life processes, particularly in the context of Business-to-Business (B2B) processes, multiple objects are involved in a process. Recently, Object-Centric Event Logs (OCELs) have been introduced to capture the information of such processes, and several process discovery techniques have been developed on top of OCELs. Yet, the output of the proposed discovery techniques on real OCELs leads to more informative but also more complex models. In this paper, we propose a clustering-based approach to cluster similar objects in OCELs to simplify the obtained process models. Using a case study of a real B2B process, we demonstrate that our approach reduces the complexity of the process models and generates coherent subsets of objects which help the end-users gain insights into the process.


\keywords{Clustering \and Object-Centric Process Mining\and Convergence.}
\end{abstract}
\section{Introduction}\label{introduction}

Process mining is a field of science bridging the gap between data-oriented analysis and process-oriented analysis, which aims to extract knowledge from event logs~\cite{van2016data}. Process mining techniques are categorized into three types: process discovery, conformance checking, and process enhancement. Process discovery extracts abstract process knowledge using visual process models. Process discovery techniques have been improved to handle complex and large event logs, e.g., Inductive Miner~\cite{van2016data}. However, the application of process discovery techniques in flexible environments such as product development leads to spaghetti process models with an overwhelming number of connections~\cite{song2008trace}. One solution is using clustering techniques to group the process instances with similar behavior. Several clustering techniques have been proposed on top of traditional event logs~\cite{boltenhagen2019generalized,bose2009context,de2021expert,di2016clustering,lehto2020discovering,luengo2011applying,seeliger2018finding,song2008trace,taub2022tracking,veiga2009understanding,yeshchenko2019comprehensive,zandkarimi2020generic}, nevertheless, in reality, multiple objects interact with each other in a process~\cite{berti2022filtering,berti2022scalable,esser2021multi,farhang2022python,waibel2022causal}, for example, considering a Purchase-to-Pay (P2P) process where \textit{orders}, \textit{items}, and \textit{customers} are involved~\cite{ghahfarokhi2021ocel,ghahfarokhi2021process}. Several process discovery techniques have been developed on top of event logs with multiple case notions~\cite{berti2018extracting,cohn2009business,lu2015discovering,meroni2018multi,meroni2017artifact,nooijen2012automatic}. For example, Object-Centric~DFGs (OC-DFGs), used throughout this paper, are one of the object-centric process models developed on top of Object-Centric Event Logs (OCELs). An OC-DFG is a Directly-Follows Graph (DFG) where relations are colored based on object types~\cite{berti2018extracting}. Several examples of such models are shown in the remainder. 
\begin{figure}[b]
\centering
\vspace{-0.5cm}
\includegraphics[width=1.03\textwidth]{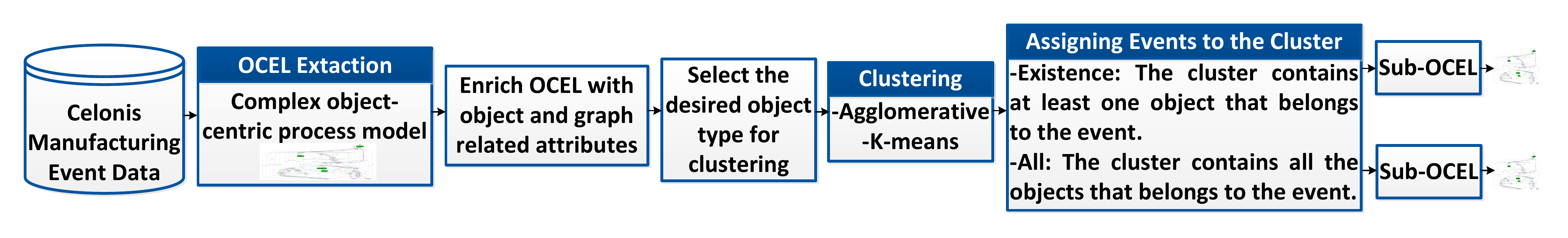}
\caption{Overview of the proposed framework.}
\label{framework}
\end{figure}
\begin{table}[t]
\caption{Informal representation of the events. Each row shows an event that has an identifier, a
timestamp, an activity, related objects, and attributes.} 
\centering 
\resizebox{1\textwidth}{!}{ 
\begin{tabular}{c c c c c c c c c } 
\hline\hline 
id &      activity &         timestamp & batch & order & customer &  net price &  gross price \\[0.5ex] 
\hline 
$e_1$ &         \emph{order~creation} & \emph{2020-04-13 11:20:01.527+01:00} &  \{\} &  \{$o_1$\} &  \{$c_1$\} & \textit{146.8}& \textit{154.8}\\ 
$e_2$ &  \emph{print~of~production~order} & \emph{2020-04-15 08:21:01.527+01:00} &        \{$b_1$, $b_2$\} &  \{$o_1$\} &                   \{$c_1$\} &          \textit{285.8}& \textit{301.3}  \\
$e_3$ &        \emph{Loading} & \emph{2020-05-09 08:22:01.527+01:00} &    \{$b_1$,$b_3$\} &  \{$o_1$\}&  \{$c_1$\}&             \textit{272.47}  & \textit{312.4}\\
... &        ... & ... &        ... &     ... &  ... &  ... &       ...            \\ [1ex] 
\hline 
\end{tabular}
}
\label{informalevents} 

\end{table}

\begin{table}[t]
\caption{Informal representation of the objects. Each row shows the properties of the objects, e.g, the treatment type for $b_1$ is painting.} 
\centering 
\resizebox{0.37\textwidth}{!}{ 
\begin{tabular}{c  c c c c c c c c c} 
\hline\hline 
id & type &  treatment & workplace \\[0.5ex] 
\hline 

$b_1$ &        \emph{batch} &   \emph{painting}&    \emph{plant~1} \\
$b_2$ &        \emph{batch} &   \emph{polishing} &    \emph{plant~1} \\
$o_1$ &       \emph{order} &        ~ &          ~  \\
... &       ... &        ... &      ... \\       
\hline 
\end{tabular}
}
\label{informalobjects} 
\end{table}

In this paper, we present a clustering-based approach, shown in Figure~\ref{framework}, which uses the relations between objects and events in clustering. First, we extracted an OCEL from a Business-to-Business (B2B) process. Then, we enriched the extracted OCEL with a few graph-related attributes, e.g., centrality measures. Afterward, we selected a clustering object type and applied data clustering algorithms to group similar objects, i.e., clusters of objects. The challenge occurs when we intend to assign events to the clusters. We propose two approaches to address this challenge:
\begin{itemize}
\item \emph{Existence}: If we directly assign events to the clusters by considering that the event should contain at least one of the objects in the cluster, then the same event may appear in several clusters. For example, consider the B2B OCEL shown in Tables~\ref{informalevents} and~\ref{informalobjects}, where \emph{customer}, \emph{order}, and \emph{batch} are the possible case notions. When we apply clustering based on \textit{batch}, then two \textit{batches} in the same event may end up in two different clusters. This results in the duplication of that event. This is due to the convergence in OCELs where an event can contain multiple objects of the same type~\cite{ghahfarokhi2021ocel}.
\item \emph{All}: In this approach, to avoid convergence, we assign an event to a cluster, if the cluster contains all the objects that exist in that event. Following this approach, we miss the event whose objects are distributed in several clusters. Consider the process shown in Tables~\ref{informalevents} and~\ref{informalobjects}, if $b_1$ and $b_2$ end up in different clusters, then we miss $e_2$, because all the batches of $e_2$ are not in the same cluster. Nevertheless, following this approach no duplication of events occurs.
\end{itemize}

To evaluate the quality of the discovered OC-DFGs, we provide initial complexity measures for OC-DFGs. Using the proposed clustering techniques and quality measures, we achieved a set of meaningful OC-DFGs with almost the same fitness but less complexity in comparison with the initial model.

The remaining part of the paper is organized as follows. Section~\ref{runningexample} presents the running example that is used throughout the paper. Then, in Section~\ref{Preliminaries}, we present some preliminary concepts that will be used throughout the paper. In Section~\ref{Object Profiles}, we discuss the object profile extraction and enrichment. Afterward, in Section~\ref{Clustering in Object-Centric Event Logs}, we describe our proposed clustering-based approach in OCELs. Then, in Section~\ref{evaluation}, we provide some experiments on the running example using our approach where we obtain simplified process models. Finally, Section~\ref{conclusion} concludes the paper and provides future work.

\section{Running Example}\label{runningexample}
\begin{figure}[b]
\centering
\vspace{-0.3cm}
\includegraphics[width=1\textwidth]{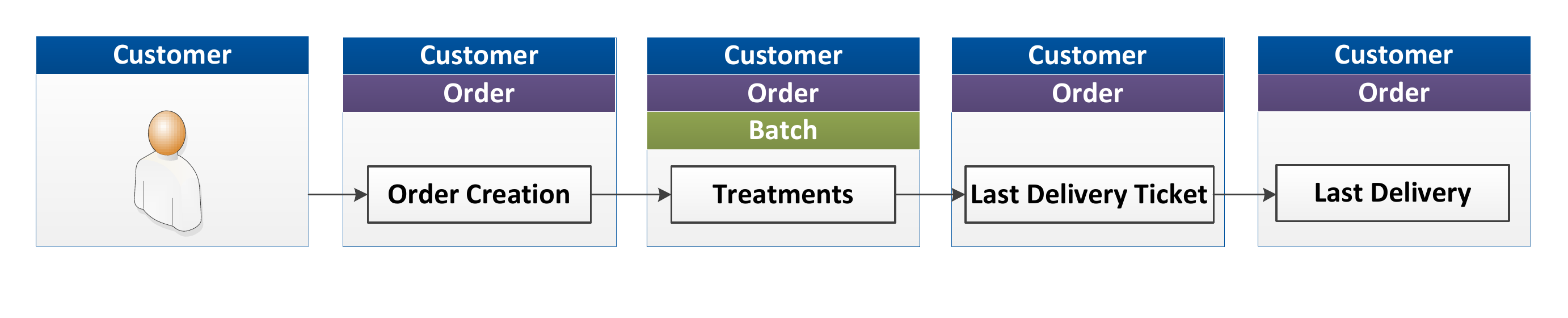}
\vspace{-0.5cm}
\caption{An abstract view of the extracted B2B process.}
\label{abstractprocess}
\vspace{-0.5cm}
\end{figure}

To evaluate our approach on real-world data, we have extracted OCEL from a B2B process, anonymized and stored in Celonic Process Analytics Workspace. The respective industry performs surface treatment services such as coating and polishing mainly for the automotive industry. Figure~\ref{abstractprocess} presents the generic process routine and the associated object types (i.e., \textit{customer}, \textit{order}, and \textit{batch}). As shown in the figure, the process starts with
the \emph{order~creation} activity. \textit{Customers} send their \textit{order} to the company and request for specific treatments. The orders will be split into \textit{batches} to fit production machines. After applying the requested treatments, respective \textit{batches} of each \textit{order} will be packed together to be shipped back to the \textit{customers}. 

To provide a comprehensive vision of the process, we have used OCEL standard by considering three possible case notions. Tables~\ref{informalevents} and~\ref{informalobjects} show the extracted OCEL where the full Table~\ref{informalevents} consists of 9004 events with different treatment activities, that are anonymized in this data. Moreover, the OC-DFG extracted from the whole process is an unreadable spaghetti model that does not give insights about the process. To derive simpler models, we can divide the OCEL into several sub-logs using clustering techniques. To apply clustering methods on objects in the OCEL, we need to extract object profiles from OCEL. In the next section, we describe the extraction of object profiles.


\section{Preliminaries} 
\label{Preliminaries}
\subsection{Object-Centric Event Logs}
\label{Object-Centric Event Logs}

First we define universes that are used throughout the paper:

\begin{definition}[Universes]
\begin{itemize}
\item ${U_e}$ is the universe of event identifiers, e.g., $\{ e_1, e_2, e_3 \} \subseteq {U_e}$

\item ${U_{act}}$ is the universe of activities, e.g., $\{ order~creation, last~delivery\} \subseteq {U_{act}}$

\item ${U_{att}}$ is the universe of attribute names, e.g., $\{gross~price, net~price\} \subseteq {U_{att}}$

\item ${U_{val}}$ is the universe of attribute values, e.g., $\{200.0, 302.0, painting\} \subseteq {U_{val}}$

\item ${U_{typ}}$ is the universe of attribute types., e.g., $\{string, integer, float\} \subseteq {U_{typ}}$

\item ${U_{o}}$ is the universe of object identifiers, e.g., $\{ o_1, b_1 \} \subseteq {U_{o}} $

\item ${U_{ot}}$ is the universe of objects types, e.g., $\{order, batch\} \subseteq {U_{ot}} $

\item ${U_{timest}}$ is the universe of timestamps, e.g., $\emph{\scriptsize 2020-04-09T08:21:01.527+01:00} \in {U_{timest}}$

\end{itemize}
\end{definition}

Using the universes above, we define object-centric event logs. 

\begin{definition}[Object-Centric Event Log]\label{def:objCentrEventLog}
An object-centric event log is a tuple $L{=}(E, AN, AV, AT, OT, O, \pi_{typ}, \pi_{act}, \pi_{time}, \pi_{vmap}, \pi_{omap}, \pi_{otyp}, \pi_{ovmap}, \leq)$ such that:
\begin{itemize}

\item $E \subseteq {U_e}$ is the set of event identifiers, e.g., $e_1$ in Table~\ref{informalevents}.

\item $AN \subseteq {U_{att}}$ is the set of attributes names, e.g., $gross~price$ in Table~\ref{informalevents}.

\item $AV \subseteq {U_{val}}$ is the set of attribute values, e.g., $plant~1$ in Table~\ref{informalobjects}.

\item $AT \subseteq {U_{typ}}$ is the set of attribute types. For example, the type of the attribute workplace in Table~\ref{informalobjects} is string.
 
\item $OT \subseteq {U_{ot}}$ is the set of object types. For example, in Table~\ref{informalobjects}, for the first object, the type is batch.

\item $O \subseteq {U_{o}}$ is the set of object identifiers, e.g.,  $o_1$ in Table~\ref{informalobjects}.

\item $\pi_{typ} : AN \cup AV \rightarrow AT$ is the function associating an attribute name or value to its corresponding type. For example, in Table~\ref{informalevents}, $\pi_{typ}(net~price) = float$.

\item $\pi_{act} : E \rightarrow {U_{act}}$ is the function associating an event to its activity, e.g., $\pi_{act}(e_1)=order~creation$ in Table~\ref{informalevents}.

\item $\pi_{time} : E \rightarrow {U_{timest}}$ is the function associating an event to a timestamp, e.g., $\pi_{time}(e_1)=$\emph{2020-04-13 11:20:01.527+01:00}  in Table~\ref{informalevents}.

\item $\pi_{vmap}: E \rightarrow (AN \not\rightarrow AV)$ is the function associating an event to its attribute value assignments, e.g., $\pi_{vmap}(e_1)(net~price)=$$146.8$ in Table~\ref{informalevents}, 

\item $\pi_{omap} : E \rightarrow \mathcal{P}(O)$ is the function associating an event to a set of related object identifiers, e.g., $\pi_{omap}(e_1) = \{ o_1, c_1\}$ in Table~\ref{informalevents}.

\item $\pi_{otyp} \in O \rightarrow OT$ assigns precisely one object type to each object identifier, e.g., $\pi_{otyp}(o_1) = order$ in Table~\ref{informalobjects}, .

\item $\pi_{ovmap} : O \rightarrow (AN \not\rightarrow AV)$ is the function associating an object to its attribute value assignments, e.g., $\pi_{ovmap}(b_1)(workplace) = $ $plan~1$ in Table~\ref{informalobjects}.

\item $\leq$ is a total order (i.e., it respects the anti-symmetry, transitivity, and connexity properties).
\end{itemize}
\end{definition}

To summarize, an event log consists of information about events and objects involved in the events. Dealing with object-centric event logs starts from log flattening. Therefore, by selecting an object type that we aim to cluster, we transform an object-centric event log into a traditional event log. 

\begin{definition}[Ot-Flattened Log]\label{def:ologFlattening} Let $L = (E, AN, AV, AT, OT, O, \pi_{typ}, \pi_{act}, \\ \pi_{time}, \pi_{vmap},\pi_{omap}, \pi_{otyp}, \pi_{ovmap}, \leq)$ be an OCEL, and $\textrm{ot} \in OT$ be an object type.
We define ot-flattened log as $FL(L, \textrm{ot}) = (E^{ot}, \pi_{act}^{ot}, \pi_{time}^{ot}, \pi_{case}^{ot}, \leq^{ot})$ where:
\begin{itemize}
\item $E^{ot} = \{ e \in E ~ \arrowvert ~ \exists_{o \in \pi_{omap}(e)} ~ \pi_{otyp}(o) = \textrm{ot}\}$,
\item $\pi_{act}^{ot} = {\pi_{act}}_{|E^{ot}}, i.e., \pi_{act}~with~the~domain~restricted~to~E^{ot}$,
\item $\pi_{time}^{ot} = {\pi_{time}}_{|E^{ot}}, i.e., \pi_{time}~with~the~domain~restricted~to~E^{ot},$
\item For $e \in E^{ot}$, $\pi_{case}^{ot}(e) = \{ o \in \pi_{omap}(e) ~ \arrowvert ~ \pi_{otyp}(o) = ot \}$, and
\item $\leq^{ot} = \{ (e_1, e_2) \in \leq ~ \arrowvert ~ \exists_{o \in O} ~ \pi_{otyp}(o) = \textrm{ot} ~ \wedge ~ o \in \pi_{omap}(e_1) \cap \pi_{omap}(e_2) \}$
\end{itemize}
\end{definition}

Using the flattened log, we extract object profiles from OCELs that will be comprehensively described in Section~\ref{Object Profiles}. To increase the number of features of objects for clustering, we enrich the OCEL with some graph-related attributes. 
Next, we describe the graph theory concepts that we used to enrich the OCELs.

\begin{definition}[Directed Graph]
A directed graph is a pair $G=(V,E)$ where~\cite{bender2010lists}:
\begin{itemize}
\item V is a set of vertices (nodes).
\item $E\subseteq \{(v_1,v_2)\in V\times V~|~v_1\not=v_2\}$ is a set of edges, which are ordered pairs of distinct vertices. In a \textit{weighted} directed graph each node is assigned to a weight through the function $f:E\rightarrow \mathbb{R}$.
\end{itemize}
\end{definition}
An example of a weighted graph is shown in Figure~\ref{graph}.
\begin{definition}[Path]
A path in graph $G=(V,E)$ is a sequence of vertices $P=\langle v_1,...,v_n \rangle\in V\times ...\times V$ such that $(v_i,v_{i+1})\in E$ for $1\leq i<n$.

$Example$: In the graph in Figure \ref{graph}, there is $P=\langle a,b,d \rangle$.
\begin{figure}[tb]
        \centering
        \vspace{-0.16cm}
        \includegraphics[width=0.53\textwidth]{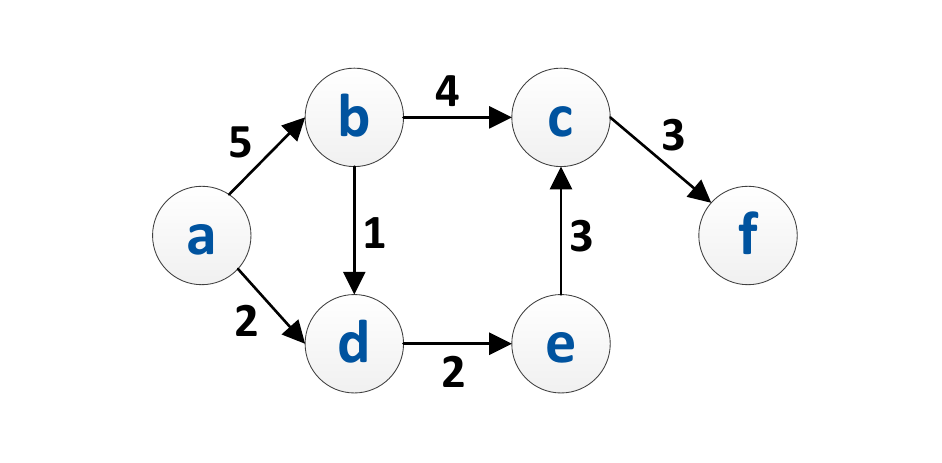}
        \vspace{-0.4cm}
        \caption{A directed graph.}
        \vspace{-0.37cm}
        \label{graph}
\end{figure}
\end{definition}

\begin{definition}[Shortest Path]
Let $G=(V,E)$ be a directed graph and $v_1, v_n \in V$. Given a weight function $f:E\rightarrow \mathbb{R}$, the shortest path from $v_1$ to $v_n$ is the path $SP=\langle v_1,...v_n \rangle$ such that over all possible paths from $v_1$ to $v_n$ it minimizes $\sum_{i=1}^{n-1} f(v_i, v_{i+1})$.

$Example$: In Figure \ref{graph}, the shortest path between node $a$ and node $c$ in weighted directed graph is $P=\langle a,d,e,c \rangle$ since $ f(a,d)+f(d,e)+f(e,c)=7$ while for example for paths $P=\langle a,b,c \rangle$ we have $f(a,b)+f(b,c)=9$.

\end{definition}




The next definitions are related to centrality. In graph theory, centrality is a number or ranking assigned to all nodes in a graph, indicating each node's position in the graph. Each type of centrality illustrates a certain type of importance or influence that a node can have.

\begin{definition}[In-Degree Centrality]
Let $G=(V,E)$ be a directed graph and $v_1\in V$. Then we define $deg_{in}(v_1)$ as the number of incoming edges to $v_1$, i.e., $deg_{in}(v_1) = |\{(v, v') \in E~|~v' = v_1\}|.$

$Example$: in Figure \ref{graph}, $deg_{in}(b)=1$ and $deg_{in}(c)=2$.

\end{definition}

\begin{definition}[Out-Degree Centrality]
Let $G=(V,E)$ be a directed graph and $v_1\in V$. Then we define $deg_{out}(v_1)$ as the number of outcoming edges from $v_1$, i.e., $deg_{out}(v_1) = |\{(v',v) \in E~|~v' = v_1\}|.$

$Example$: in Figure \ref{graph}, $deg_{out}(a)=2$ and $deg_{out}(b)=2$.
\end{definition}

\begin{definition}[Closeness Centrality]
Let $G=(V,E)$ be a directed graph and $v\in V$. Then we define closeness centrality of $v$ as the reciprocal of the sum of the length of the shortest paths between $v$ and all other nodes in the graph. Normalized closeness centrality is defined as:
\begin{equation}
    C_C(v)= \frac{|V|-1}{\sum_{y\in V} {SP(v,y)}}
\end{equation}
Where $SP(v,y)$ is the shortest path between vertices $v$ and $y$. Therefore, the more central a node is, the closer it is to all other nodes.
\end{definition}

\begin{definition}[Harmonic Centrality]
Let $G=(V,E)$ be a directed graph and $v\in V$. Harmonic centrality is defined as:
\begin{equation}
    C_H(v)= \sum_{y \in V\setminus\{v\}} {\frac{|V|-1}{SP(v,y)}}
\end{equation}
where $SH(y,x)$ is the shortest path between vertices $v$ and $y$.
\end{definition}

Using described graph-related attributes, we enrich the object information. In the next section, we describe how we comprise object profiles and enrich them using the graph-related features to apply clustering techniques.

\section{Object Profiles}\label{Object Profiles}
Clustering algorithms group sets of similar points. In clustering objects of the OCEL, the points to be clustered are object profiles. To start clustering, we preprocess the data and enrich it with some additional features for each object. Below, we describe how we enrich the object attributes with graph attributes. First, we extract the related trace to an object using the flattened log.

\begin{definition}[Trace] Given an ot-flattened log $FL = (E^{ot}, \pi_{act}^{ot}, \pi_{time}^{ot}, \pi_{case}^{ot},\\ \leq^{ot})$, we define the following operations:

\begin{itemize}
\item $\pi^{ot}_{act}(FL) = \{ \pi^{ot}_{act}(e) ~ \arrowvert ~ e \in E^{ot} \}$
\item $\pi^{ot}_{case}(FL) = \cup_{e \in E^{ot}} ~ \pi^{ot}_{case}(e)$
\item For $c \in \pi^{ot}_{case}(FL)$, $case^{ot}_{FL}(c) = \langle e_1, \ldots, e_n \rangle$ where:
\begin{itemize}
\item $\{ e_1, \ldots, e_n \} = \{ e \in E^{ot} ~ \arrowvert ~ c \in \pi^{ot}_{case}(e) \}$
\item $\forall_{1 \leq i < n} ~ e_i < e_{i+1}$
\end{itemize}
\item Given $c \in \pi^{ot}_{case}(FL)$ and $case^{ot}_{FL}(c) = \langle e_1, \ldots, e_n \rangle$,we define \\ $\textrm{trace}_{FL}(c) = \langle \pi^{ot}_{act}(e_1), \ldots, \pi^{ot}_{act}(e_n) \rangle$
\end{itemize}

\end{definition}

Moreover, to provide derived attributes, we create a directed weighted graph based on the sequence of activities of each object.

\begin{definition}[Trace Graph]
Let $FL = (E^{ot}, \pi_{act}^{ot}, \pi_{time}^{ot}, \pi_{case}^{ot}, \leq^{ot})$ be a flattened OCEL  and $c \in \pi^{ot}_{case}(FL)$ be an object. For the object trace $\textrm{trace}_{FL}(c) = \langle a_1, \ldots, a_n \rangle$, we define the corresponding directed weighted graph of the trace as: 

$G_{\textrm{trace}_{FL}(c)}=(V,E)$ with the weight function $\pi^{ot}_{freq}:E\rightarrow \mathbb{R}$ where:

\begin{itemize}
    \item $V=\{a_1, ..., a_n\}$
    \item $E=\{(a_i,a_{i+1})|1\leq i<n\}$
    \item For $(x,y)\in E$, $\pi^{ot}_{freq}(x,y)=|\{(a_1, a_2)\in E~|~(a_1, a_2)=(x,y)\}|$
\end{itemize}

\begin{figure}[b]
\centering
\vspace{-0.5cm}
\includegraphics[scale=0.59]{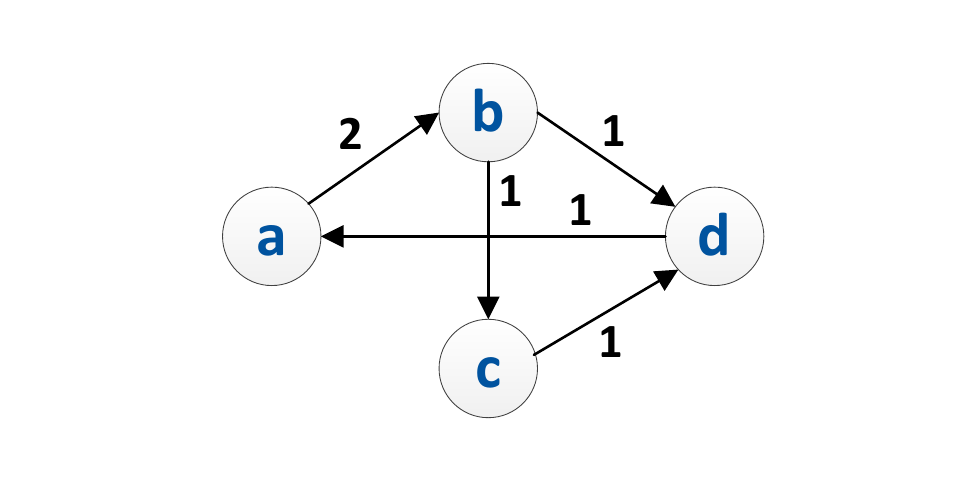}
\vspace{-0.4cm}
\caption{The graph of the trace $\textrm{trace}_{FL}(c)$.}
\label{fig:trace}
\end{figure}
\end{definition}

The graph for trace $\sigma = \langle a,b,c,d,a,b,d \rangle$ is presented in Figure \ref{fig:trace}. For each object we calculate the trace graph and for each node in every graph we find \emph{in-degree centrality}, \emph{out-degree centrality}, \emph{closeness centrality}, and \emph{harmonic centrality}, described in Section~\ref{Preliminaries}. As an illustration, in Figure~\ref{fig:trace} the node list is $V=\{a,b,c,d\}$ and the corresponding \textit{in-degree centrality} vector is $(1,1,1,2)$. However, we need to assign a unique value to this object as the \textit{in-degree centrality}. Thus, for each trace graph, the mean, variance, and standard deviation of all vector elements are calculated and inserted in the object attribute. For the mentioned \textit{in-degree centrality} vector (i.e., $(1,1,1,2)$) the mean is 1.25, the variance is 0.25, and the standard deviation is 0.5. These values are added to the related object attributes as \textit{in-degree centrality means}, \textit{in-degree centrality variance}, and \textit{in-degree centrality standard deviation}. For other features such as \textit{closeness centrality}, we follow the same procedure. Using all these features, we enrich the object attributes with graph related attributes.
\begin{table}[t]
\vspace{-0.2cm}
\caption{Object profiles extracted from an OCEL.}
\centering
\resizebox{1\textwidth}{!}{ 
\begin{tabular}{|c|c|c|c|c|c|c|c|}
\hline
object ID & trace                                  & treatment & workplace   & ... & \begin{tabular}[c]{@{}c@{}}in-degree \\ centrality mean\end{tabular} & \begin{tabular}[c]{@{}c@{}}in-degree\\ centrality std\end{tabular} & \begin{tabular}[c]{@{}c@{}}in-degree\\ centrality var\end{tabular} \\ \hline
$b_1$      & $\langle print~of~production~order, loading \rangle$          & $painting$  & $plan~1$ & ... & 0.50                                                                 & 0.50                                                                & 0.25                                                               \\ \hline
$b_2$      & $\langle print~of~production~order, ..., lubricate\rangle$ & $polishing$ & $plan~1$ & ... & 1.00                                                                & 0.00                                                               & 0.00                                                              \\ \hline
$b_3$      & $\langle loading, painting \rangle$        & $painting$  & $plan~2$ & ... & 0.50                                                                 & 0.50                                                               & 0.25                                                               \\ \hline
\end{tabular}}
\label{objectprofiles}
\vspace{-0.4cm}
\end{table}
Now, using object attributes and object trace, we define object profile which is used as an input for clustering.

\begin{definition}[Object Profile]\label{object profile} Let $FL = (E^{ot}, \pi_{act}^{ot}, \pi_{time}^{ot}, \pi_{case}^{ot}, \leq^{ot})$ be a flattened OCEL. We define object profile function for $o\in O$ where $\pi_{otyp}(o)=ot$ as $op:U_{o}{\rightarrow}~U^*_{act} \times U_{val} \times ... \times U_{val}$ such that $op(o)=(trace_{FL}(o), \pi_{ovmap}(o)(att_1),\\ ..., \pi_{ovmap}(o)(att_n))$ and $att_1, ..., att_n \in dom(\pi_{ovmap}(o)).$
\end{definition}

An example of the extracted profiles is shown in Table~\ref{objectprofiles} where the batch profiles are represented. For example for $b_1$, the extracted trace, $treatment$, $workplace$, and $in{-}degree~centrality~mean$ are the object attributes that constitute the profile for $b_1$. To sum up, using the graph features, we enrich the object profiles and the output of the profile extraction step is the enriched profiles. Based on this information, we apply clustering methods to the objects.

\section{Clustering in Object-Centric Event Logs}\label{Clustering in Object-Centric Event Logs}
In this section, we present the clustering algorithm. First, we describe the distance measures that we used to find the similarity between object profiles. Afterward, we describe the two clustering techniques that we used in this research.

\subsection{Distance Measures}
\label{Distance Measures}
Clustering results are affected by the distance measures that are used to measure the distance between object profiles. An example of object profiles is shown in Table~\ref{objectprofiles}. As the table illustrates, an object profile consists of the object's control flow and the attribute values which can be numerical or categorical. Therefore, different distance measures are needed to calculate the distance between object profiles. To calculate the distance between the attributes related to the control flow, numerical attributes, and categorical attributes we apply Levenshtein, Euclidean, and String Boolean distance that are described below, respectively.

We used Levenshtein distance to measure the distance between two sets of activities where we should transform one set of activities to another set of activities. Therefore, a set of operations that are substitution, insertion, and deletion are needed to be done on one of the sequences. The mathematical representation of these edit distances is described in~\cite{bose2009context}. 
Euclidean Distance is used to measure the distance between numerical values and String Boolean Distance is used to measure the distance between categorical values. If the categorical values are the same the distance is zero otherwise the distance is one.

Using described distance metrics, we find the distance of the objects from each other to apply clustering algorithms. In the next section, we describe the clustering algorithms that we utilized in this paper.

\subsection{Clustering Algorithm}
\label{Clustering Algorithm}
In this section, we shortly explain clustering algorithms, i.e., K-means and Agglomerative Clustering. These clustering algorithms can be applied on the object profiles, described in Section~\ref{Object Profiles}, to create clusters of homogeneous objects.
\begin{itemize}
    \item K-means Clustering: K-means technique is one of the most common clustering methods among partitioning methods. K-means algorithm clusters data into k clusters based on minimizing within-cluster sum-of-squares criteria.
    \item Hierarchical Clustering: Hierarchical clustering is used to cluster data based on a hierarchy of clusters. There are main approaches in the hierarchical clustering method: agglomerative (i.e., a bottom-up approach) and divisive (i.e., a top-down approach). In this paper, we have applied agglomerative clustering where generates clusters through merging the nearest objects, i.e., smaller clusters of objects are combined to make a larger cluster. 
\end{itemize}
Using each of the clustering techniques above, we map a set of objects with the same type (e.g., batch) onto the clusters containing a set of objects:
\begin{definition}[Clustering]\label{def:OCEL clustering}
Let $L= (E, AN, AV, AT, OT, O, \pi_{typ}, \pi_{act}, \pi_{time},\\ \pi_{vmap}, \pi_{omap}, \pi_{otyp}, \pi_{ovmap}, \leq)$ be an OCEL and $ot$ be an object type which we aim to do clustering for. Clustering is a function $cl: \mathcal{P}({U_{o}}) {\rightarrow} \mathcal{P}{(\mathcal{P}({U_o})})$ such that $cl(O)=\left \{O_1, O_2, ..., O_n \right \}$ where $\forall o \in O ~ \pi_{otyp}(o)=ot$ and $\bigcup_{i=1}^{n}(O_i)=O$.
\end{definition}

By applying clustering methods on objects and using their profiles, described in Definition~\ref{object profile}, we obtain clusters of objects with the same type. In the next section, we describe how we transform the results of clustering into an OCEL.
\subsection{Transformation of the Clustering Results into OCEL}
\label{Transformation of the Clustering Result into OCEL}
To extract process models from the obtained clusters, we should assign the clusters to the corresponding events. Here, we propose two approaches based on the state of the objects in the event. 
\begin{itemize}
    \item \emph{Existence}: In this approach, we assign an event to the cluster, containing at least one object existing in that event. This approach leads to duplication in events that is described in Section~\ref{introduction}. An example is shown in Figure~\ref{existence} where $e_3$ is in both clusters, since $b_1$ and $b_3$ are in separate clusters. Now, we formalize the notion of \emph{existence} approach as a function:
    \begin{definition}[Existence]\label{Existence}
    Let $L= (E, AN, AV, AT, OT, O, \pi_{typ}, \pi_{act},\\ \pi_{time}, \pi_{vmap}, \pi_{omap}, \pi_{otyp}, \pi_{ovmap}, \leq)$ be an OCEL and $OCL$ be a set of objects in a cluster where $OCL\in cl(O)$. Existence is a function $ex: U_{o} {\rightarrow} U_{e}$ such that $ex(OCL)=\left \{e\in E~|~\pi_{omap}(e)\cap OCL\neq \emptyset \right \}$.
    \end{definition}
    \begin{figure}[tb]
     \centering
     \vspace{-1.4cm}
     \includegraphics[scale=0.50]{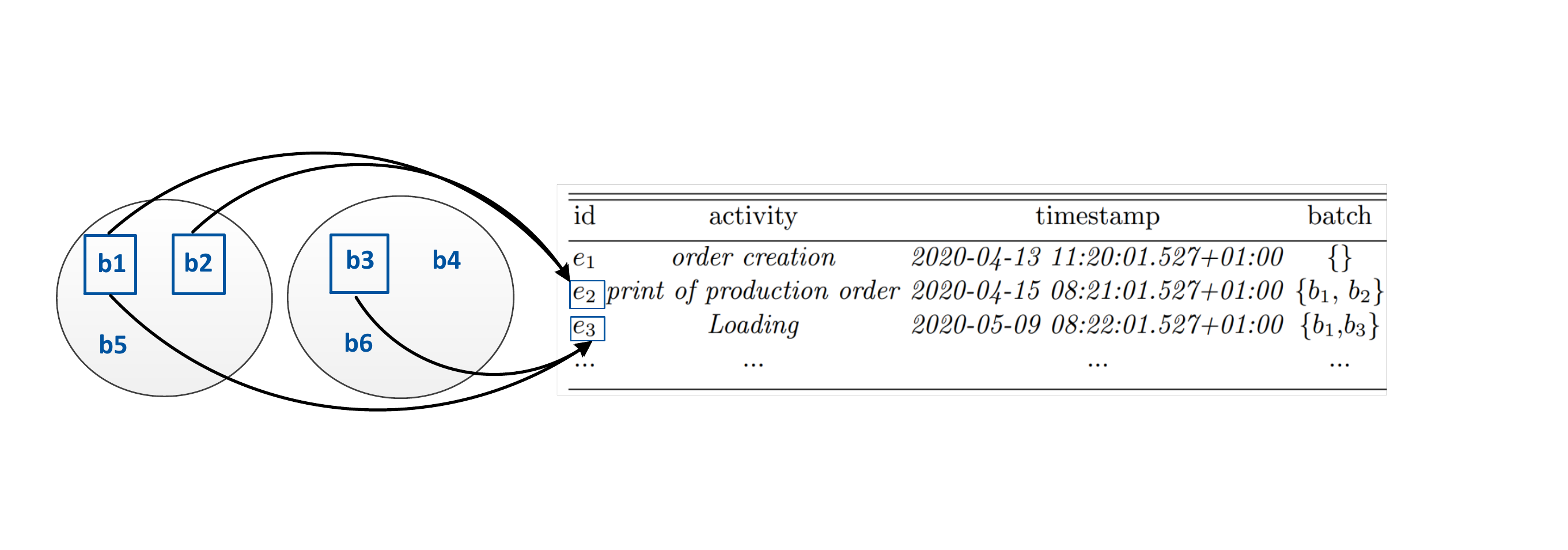}
     \vspace{-1.7cm}
     \caption{Matching clusters with the events using \emph{existence} approach.}
     \vspace{-0.5cm}
     \label{existence}
    \end{figure}
    \item \emph{All}: Assuming we do clustering based on the objects with type $ot$, in this approach, we assign an event to the cluster that contains all objects with the type $ot$ that exist in that event. This may lead to the loss of some events that can not be assigned to any of the clusters. In fact, there is no cluster that contains all the objects of the type $ot$, existing in that event. An example is shown in Figure~\ref{all} where $e_3$ is missed, since $b_1$ and $b_3$ which are the batches involved in $e_3$ are in different clusters. Here, we formalize the \emph{all} approach:
    \begin{definition}[All]\label{def:All}
    Let $L= (E, AN, AV, AT, OT, O, \pi_{typ}, \pi_{act}, \pi_{time},\\ \pi_{vmap}, \pi_{omap}, \pi_{otyp}, \pi_{ovmap}, \leq)$ be an OCEL and $OCL$ be a set of objects in a cluster where $OCL\in cl(O)$. All is a function $all: U_{o} {\rightarrow} U_{e}$ where $all(OCL)=\left \{e\in E~|~\forall o\in \pi_{omap}(e)~\pi_{otyp}(o)=ot \wedge o\in OCL \right \}$.
    \end{definition}
        \begin{figure}[tb]
     \centering
     \vspace{-0.75cm}
     \includegraphics[scale=0.50]{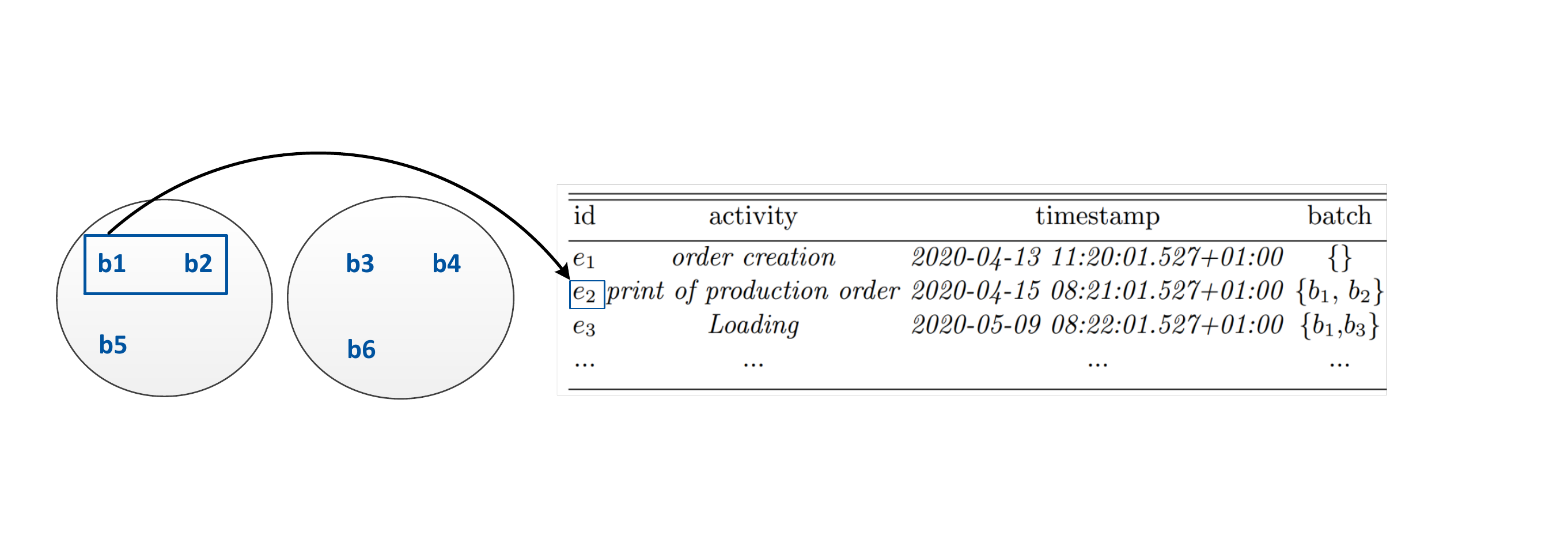}
     \vspace{-1.7cm}
     \caption{Matching clusters with the events using \emph{all} approach.}
     \label{all}
     \vspace{-0.35cm}
    \end{figure}
\end{itemize}

By applying the proposed clustering technique on OCELs, we obtain sub-logs for each cluster. However, the aim of this paper is to apply clustering techniques to obtain less complex models. Thus, in the following section, we define fitness, size, density, and some comparative complexity measures on top of OC-DFGs.

\subsection{Quality Metrics}\label{Quality Metrics}
To measure the quality of obtained models and compare them with the main model, we should define the quality metrics such as complexity. We first define the discovery of an OC-DFG which is the basis of the rest of the definitions. 

\begin{definition}[Discovery of an OCDFG]\label{def:objCentrDfg} Let $L= (E, AN, AV, AT, OT, O,\\ \pi_{typ}, \pi_{act}, \pi_{time}, \pi_{vmap}, \pi_{omap}, \pi_{otyp}, \pi_{ovmap}, \leq)$ be an OCEL. Then we define\\ $OCDFG(L)=(A, OT, F, \pi_{freqn}, \pi_{freq})$ where:
\begin{itemize}
\item $A \subseteq {U_{act}}$ is the set of activities.
\item $OT \subseteq {U_{ot}}$ is the set of object types.
\item $F \subseteq \left ( ( \{ \triangleright \} \cup A \right ) \times \left ( A \cup \{ \square \} ) \right ) \times OT$ is the set of (typed) edges.
\item $\pi_{freqn} : A \not\rightarrow \mathbb{N}$ is a frequency measure on the nodes.
\item $\pi_{freq} : F \not\rightarrow \mathbb{N}$ is a frequency measure on the (typed) edges.
\end{itemize}
\end{definition}
OC-DFGs are one of the state-of-the-art object-centric models where each object type is shown with a specific color. To evaluate the quality of OC-DFGs, we use the fitness criteria described in~\cite{berti2018extracting}. However, we also define other measures to find the complexity of the model. Next, we define the size and density of the model. The smaller the graph, the simpler the structure. 

\begin{definition}[Size]\label{def:size}
 Given an $OCDFG=(A,OT,F,\pi_{freqn},\pi_{freq})$, we define the size of the $OCDFG$ as $size(OCDFG)=|A|\times |F|$.
\end{definition}

To measure the density of the process model, we have employed the density measure of a graph introduced in \cite{lawler2001combinatorial}. The interpretation of the formula in Definition \ref{def:density} is that the more dense the graph, the more complex the model is.

\begin{definition}[Density]\label{def:density}  Given an $OCDFG=(A,OT,F,\pi_{freqn},\pi_{freq})$, we define the density of the $OCDFG$ as $density(OCDFG)=|A| / |F|$.
\end{definition}

The size and density capture the general information regarding complexity in process models, however, to evaluate our approach we should compare the complexity of the obtained process models from clusters with the main process model. Therefore, we define the concepts related to improvements in size and density. These measures are designed to compare the weighted average of the size or density of all clusters with the size or density of the main process model.

\begin{definition}[Improvement in Size Complexity]
Let $\left \{O_1, ..., O_n \right \}$ be the set of clusters obtained from $L$, and $\left \{L_1, ..., L_n\right \}$ be the set of events assigned to each cluster by applying existence or all approach, e.g., $L_1=ex(O_1), ..., L_n=ex(O_n)$. We define \textit{size complexity improvement} $C_sI$ as
\[C_sI=\frac{size(OCDFG(L))}{\frac{\sum_{i=1}^{n}|O_i|size(OCDFG(L_i)}{\sum_{i=1}^{n}|O_i|}}\]
\end{definition}

\begin{definition}[Improvement in Density Complexity]
Let $\left \{O_1, ..., O_n \right \}$ be the set of clusters obtained from $L$, and $\left \{L_1, ..., L_n\right \}$ be the set of events assigned to each cluster by applying existence or all approach, e.g., $L_1=ex(O_1), ..., L_n=ex(O_n)$. We define \textit{size complexity improvement} $C_dI$ as
\[C_dI=\frac{density(OCDFG(L))}{\frac{\sum_{i=1}^{n}|O_i|density(OCDFG(L_i)}{\sum_{i=1}^{n}|O_i|}}\]
\end{definition}

For the last two metrics, the values less than one mean that we obtained more complex models, and the values greater than one indicate that less complex models are achieved. In the next section, we evaluate our approach on a real B2B process using the described evaluation metrics.

\section{Evaluation}\label{evaluation}
To validate the proposed approach for object clustering in OCELs, we have performed a case study using the B2B dataset described in Section~\ref{runningexample} representing a treatment process. This dataset contains 9004 events and three object types, namely \emph{customer}, \emph{order}, and \emph{batch}. An \emph{order} stands for a specific treatment to be applied on number of \emph{batches} sent by a \emph{customer}. The behavior of \emph{customer} and \emph{order} are similar, i.e., each \emph{order} belongs to only one \emph{customer}. Therefore, we evaluated our approach using \emph{order} and \emph{batch}.

Figure~\ref{tab:mainmodel} shows the process model of the whole OCEL which is a spaghetti-like model and too complex to interpret. This process model is shown to the domain expert and he failed to recognize the real process behind it. Therefore, we applied the proposed clustering technique, described in Section~\ref{Clustering in Object-Centric Event Logs}, to discover simplified process models for each cluster. To find the optimal number of clusters we have employed Calinski-Harabasz, and Dendrogram for K-means and hierarchical clustering, respectively. The results confirm that at \emph{batch}-level, three or four clusters and at \emph{order}-level, two or three clusters are the best choices. Considering the optimal number of clusters, we have applied Agglomerative and K-means clustering techniques to find the clusters of objects. Both techniques were effective, nevertheless, the results of the K-means algorithm are more promising. By applying K-means clustering on the set of object profiles, we got a set of objects in each cluster. Afterward, using \emph{existence} and \emph{all} approaches we managed to assign events to the clusters. Tables~\ref{tab:kmeansexistence} and~\ref{tab:kmeansall} report the complexity and fitness of the respective models of the resulted clusters. We evaluated the obtained process models using the fitness and complexity criteria described in Section~\ref{Quality Metrics}. As the results show, the complexity of the obtained process models is reduced with the same or higher fitness. For example, the result of clustering based on batch with four clusters and using \emph{all} approach is shown in Figure~\ref{batch4}. 

\begin{table}[b]
\vspace{-0.7cm}
\caption{Some characterizations of the main model.}
\centering
\resizebox{0.55\textwidth}{!}{ 
\begin{tabular}{|c|c|c|c|c|}
\hline
\multicolumn{5}{|c|}{\textbf{The Main Model Properties}} \\ \hline
No. of Nodes & No. of Edges & Fitness & Size & Density \\ \hline
25 & 118 & 0.83 & 2950 & 4.76 \\ \hline
\end{tabular}}
\label{tab:mainmodel}
\end{table}

\begin{table}[t]
\centering
\vspace{-0.3cm}
\caption{The clustering result using K-means and \emph{existence} approach.}
\label{tab:kmeansexistence}
\resizebox{0.87\linewidth}{!}{
\begin{tabular}{|c|c|c|c|c|c|c|c|c|c|} 
\hline
\multicolumn{10}{|c|}{\begin{tabular}[c]{@{}c@{}}K-means\end{tabular}}                                                                                                                                                                                                                                  \\ 
\hline
\multicolumn{1}{|l|}{Objects} & No. of Clusters    & \multicolumn{1}{l|}{No. of Nodes} & \multicolumn{1}{l|}{No. of Edges} & \multicolumn{1}{l|}{Fitness} & \multicolumn{1}{l|}{Size} & \multicolumn{1}{l|}{Density} & Avg. Fitness          & \multicolumn{1}{l|}{$C_sI$} & \multicolumn{1}{l|}{$C_dI$}  \\ 
\hline
\multirow{5}{*}{Order}        & \multirow{2}{*}{2} & 24                                & 106                               & 0.85                         & 2544                      & 4.42                         & \multirow{2}{*}{0.85} & \multirow{2}{*}{1.22}       & \multirow{2}{*}{7.31}        \\ 
\cline{3-7}
                              &                    & 12                                & 34                                & 0.89                         & 408                       & 2.83                         &                       &                             &                              \\ 
\cline{2-10}
                              & \multirow{3}{*}{3} & 24                                & 106                               & 0.85                         & 2544                      & 4.42                         & \multirow{3}{*}{0.85} & \multirow{3}{*}{1.23}       & \multirow{3}{*}{6.69}        \\ 
\cline{3-7}
                              &                    & 12                                & 27                                & 0.86                         & 324                       & 2.25                         &                       &                             &                              \\ 
\cline{3-7}
                              &                    & 10                                & 20                                & 0.98                         & 200                       & 2                            &                       &                             &                              \\ 
\hline
\multirow{7}{*}{Batch}        & \multirow{3}{*}{3} & 20                                & 80                                & 0.81                         & 1600                      & 4                            & \multirow{3}{*}{0.84} & \multirow{3}{*}{2.34}       & \multirow{3}{*}{43.75}       \\ 
\cline{3-7}
                              &                    & 17                                & 48                                & 0.87                         & 816                       & 2.82                         &                       &                             &                              \\ 
\cline{3-7}
                              &                    & 9                                 & 20                                & 0.96                         & 180                       & 2.22                         &                       &                             &                              \\ 
\cline{2-10}
                              & \multirow{4}{*}{4} & 19                                & 75                                & 0.85                         & 1425                      & 3.95                         & \multirow{4}{*}{0.87} & \multirow{4}{*}{3.19}       & \multirow{4}{*}{44.61}       \\ 
\cline{3-7}
                              &                    & 9                                 & 20                                & 0.96                         & 180                       & 2.22                         &                       &                             &                              \\ 
\cline{3-7}
                              &                    & 7                                 & 16                                & 0.87                         & 112                       & 2.29                         &                       &                             &                              \\ 
\cline{3-7}
                              &                    & 11                                & 34                                & 0.89                         & 374                       & 3.09                         &                       &                             &                              \\
\hline
\end{tabular}}
\end{table}

\begin{table}[!]
\centering
\caption{The clustering result using K-means and \emph{all} approach.}
\label{tab:kmeansall}
\resizebox{0.87\linewidth}{!}{
\begin{tabular}{|c|c|c|c|c|c|c|c|c|c|} 
\hline
\multicolumn{10}{|c|}{\begin{tabular}[c]{@{}c@{}}K-means\end{tabular}}                                                                                                                                                                                                                                  \\ 
\hline
\multicolumn{1}{|l|}{Objects} & No. of Clusters    & \multicolumn{1}{l|}{No. of Nodes} & \multicolumn{1}{l|}{No. of Edges} & \multicolumn{1}{l|}{Fitness} & \multicolumn{1}{l|}{Size} & \multicolumn{1}{l|}{Density} & Avg. Fitness          & \multicolumn{1}{l|}{$C_sI$} & \multicolumn{1}{l|}{$C_dI$}  \\ 
\hline
\multirow{5}{*}{Order}        & \multirow{2}{*}{2} & 24                                & 106                               & 0.85                         & 2544                      & 4.42                         & \multirow{2}{*}{0.85} & \multirow{2}{*}{1.22}       & \multirow{2}{*}{7.31}        \\ 
\cline{3-7}
                              &                    & 12                                & 34                                & 0.89                         & 408                       & 2.83                         &                       &                             &                              \\ 
\cline{2-10}
                              & \multirow{3}{*}{3} & 24                                & 106                               & 0.85                         & 2544                      & 4.42                         & \multirow{3}{*}{0.85} & \multirow{3}{*}{1.23}       & \multirow{3}{*}{6.69}        \\ 
\cline{3-7}
                              &                    & 12                                & 27                                & 0.86                         & 324                       & 2.25                         &                       &                             &                              \\ 
\cline{3-7}
                              &                    & 10                                & 20                                & 0.98                         & 200                       & 2                            &                       &                             &                              \\ 
\hline
\multirow{7}{*}{Batch}        & \multirow{3}{*}{3} & 20                                & 78                                & 0.78                         & 1560                      & 3.9                          & \multirow{3}{*}{0.82} & \multirow{3}{*}{2.38}       & \multirow{3}{*}{43.74}       \\ 
\cline{3-7}
                              &                    & 17                                & 49                                & 0.88                         & 833                       & 2.88                         &                       &                             &                              \\ 
\cline{3-7}
                              &                    & 9                                 & 21                                & 0.98                         & 189                       & 2.33                         &                       &                             &                              \\ 
\cline{2-10}
                              & \multirow{4}{*}{4} & 19                                & 73                                & 0.86                         & 1387                      & 3.84                         & \multirow{4}{*}{0.88} & \multirow{4}{*}{3.27}       & \multirow{4}{*}{44.95}       \\ 
\cline{3-7}
                              &                    & 7                                 & 16                                & 0.87                         & 112                       & 2.29                         &                       &                             &                              \\ 
\cline{3-7}
                              &                    & 9                                 & 21                                & 0.98                         & 189                       & 2.33                         &                       &                             &                              \\ 
\cline{3-7}
                              &                    & 11                                & 34                                & 0.91                         & 374                       & 3.09                         &                       &                             &                              \\
\hline
\end{tabular}}

\end{table}

Besides the simplification of process models the discovered process models per cluster show some interesting points:
\begin {itemize}
\item In three clusters (i.e., Cluster~1, Cluster~2, and Cluster~3) the process has started with \emph{order~creation}, however, in Cluster~4 there is no \emph{order~creation}. After discussion with the expert, we realized that Cluster~4 shows the rework process of the items that experienced failures in their previous treatment process. Therefore, no \emph{order creation} is executed in these processes.
\item There is a difference between Cluster~3 and two other clusters (i.e, Cluster~1 and Cluster~2). \emph{Print of order production} is followed by \emph{hanging pieces} in Cluster~3 whereas it is followed by \emph{loading}, in Cluster~1 and Cluster~2. We recognized that the process, shown in Cluster~3, refers to small items such as nuts and bolts. Therefore, we hang them to plate both sides of them. However, cluster~1 and cluster~2 represent the process of larger items such as bottles that we should load to do the treatment.
\item  Cluster~1 and Cluster~2 illustrate the process of two different types of items since the activities that are executed between \emph{loading} and \emph{unloading} are not the same. For example, \emph{oil removing} is executed in Cluster~2 while \emph{golden layer} and \emph{surface securing} are the common activities in Cluster~1. 
\item  The \emph{last delivery ticket} activity shown in Cluster~1 and Cluster~2 shows the delivery status. When an employee finishes an order which is usually divided into several batches, the shipping process starts. Each delivery in the shipping process requires a delivery ticket. The \emph{Last delivery ticket} refers to the last shipment of an order and its respective delivery ticket.
\end{itemize}

As we see the proposed technique can distinguish different processes that exist in the initial OCEL. To sum up, we have applied the proposed clustering technique on a B2B process where multiple object types are involved. The initial process model is complex to interpret, however, to derive simpler models, we divided the OCEL into several sub-logs using the proposed clustering techniques. The obtained process models are simplified and meaningful process models that can separate different processes and help the user gain insights into the process\footnote{\scriptsize All the experiments are implemented in PM4PY~\cite{berti2019process}.}.

\begin{figure}[!]
\vspace{-0.285cm}
  \centering
    \begin{subfigure}[b]{0.93\linewidth}
    \includegraphics[width=\linewidth]{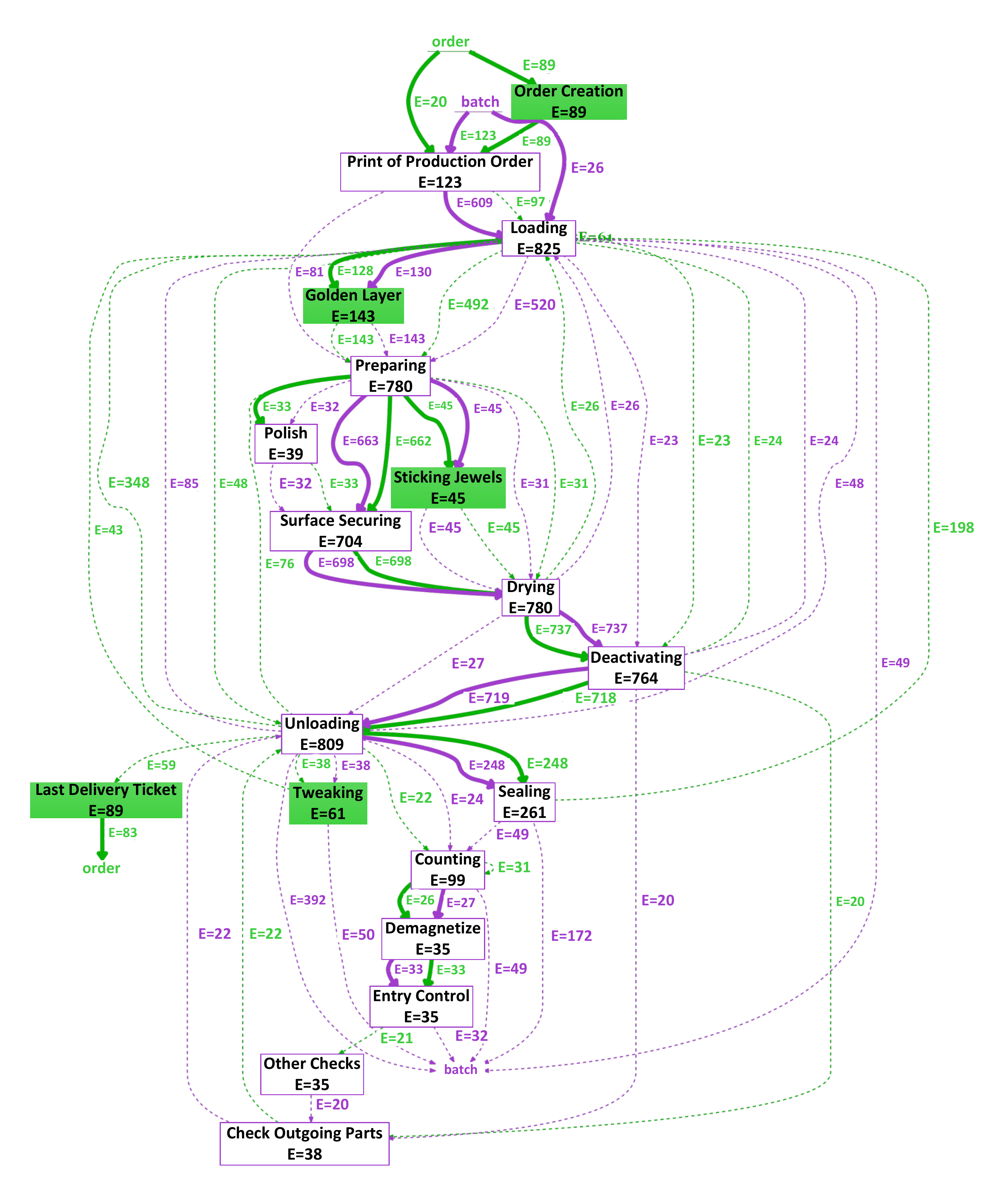}
    \vspace{-0.89cm}
    \caption{Cluster~1.}
  \end{subfigure}
  
  \begin{subfigure}[b]{0.63\linewidth}
  \vspace{-0.3cm}
    \includegraphics[width=\linewidth]{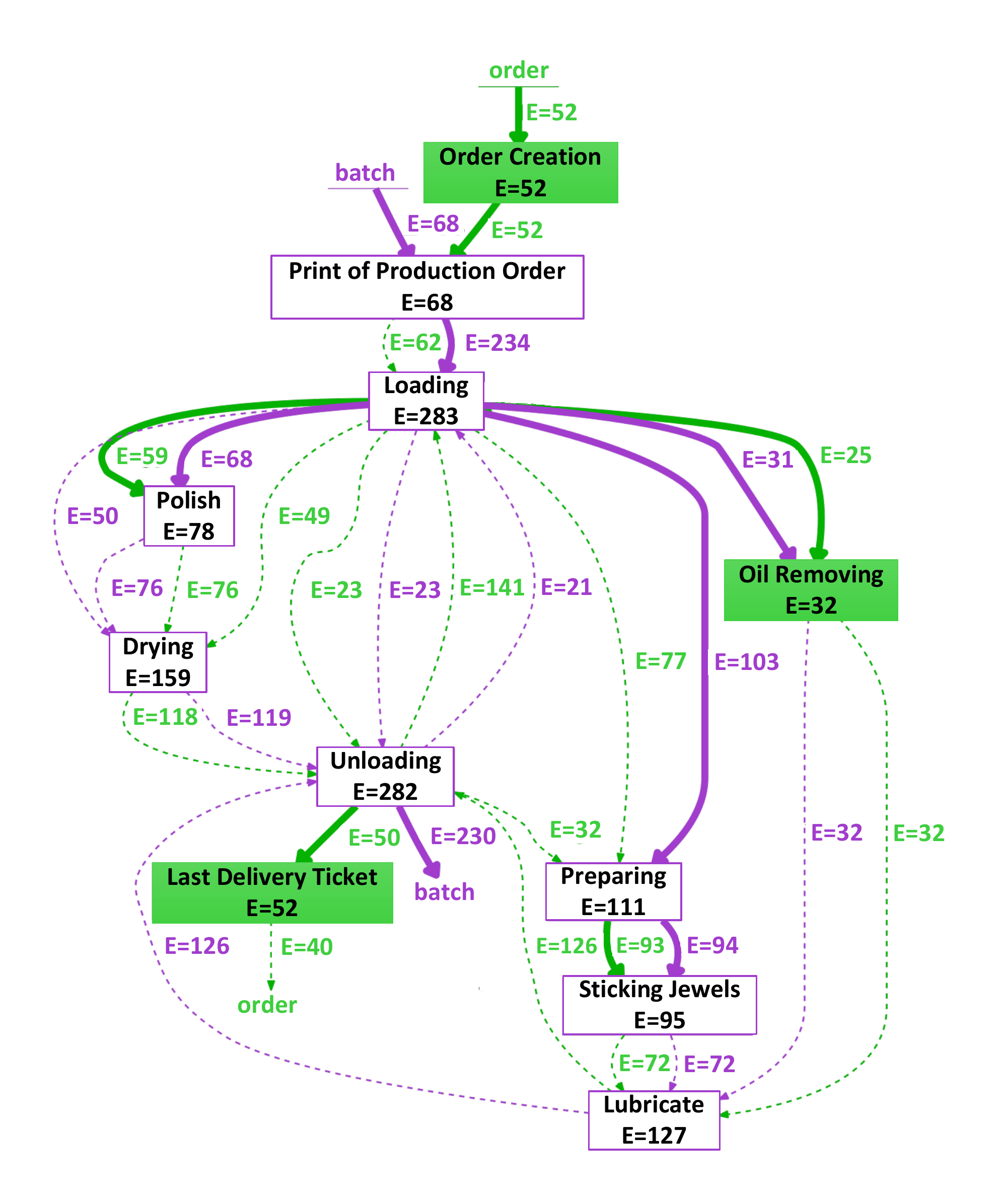}
    \vspace{-0.86cm}
    \caption{Cluster~2.}
  \end{subfigure}
    \vspace{-0.14cm}
    \caption{Final result of the clustering of batch based on \emph{all} approach.}
  \label{batch4}
\end{figure}
\begin{figure}[t]
  \ContinuedFloat
  \begin{subfigure}[b]{0.5\linewidth}
    \includegraphics[width=\linewidth]{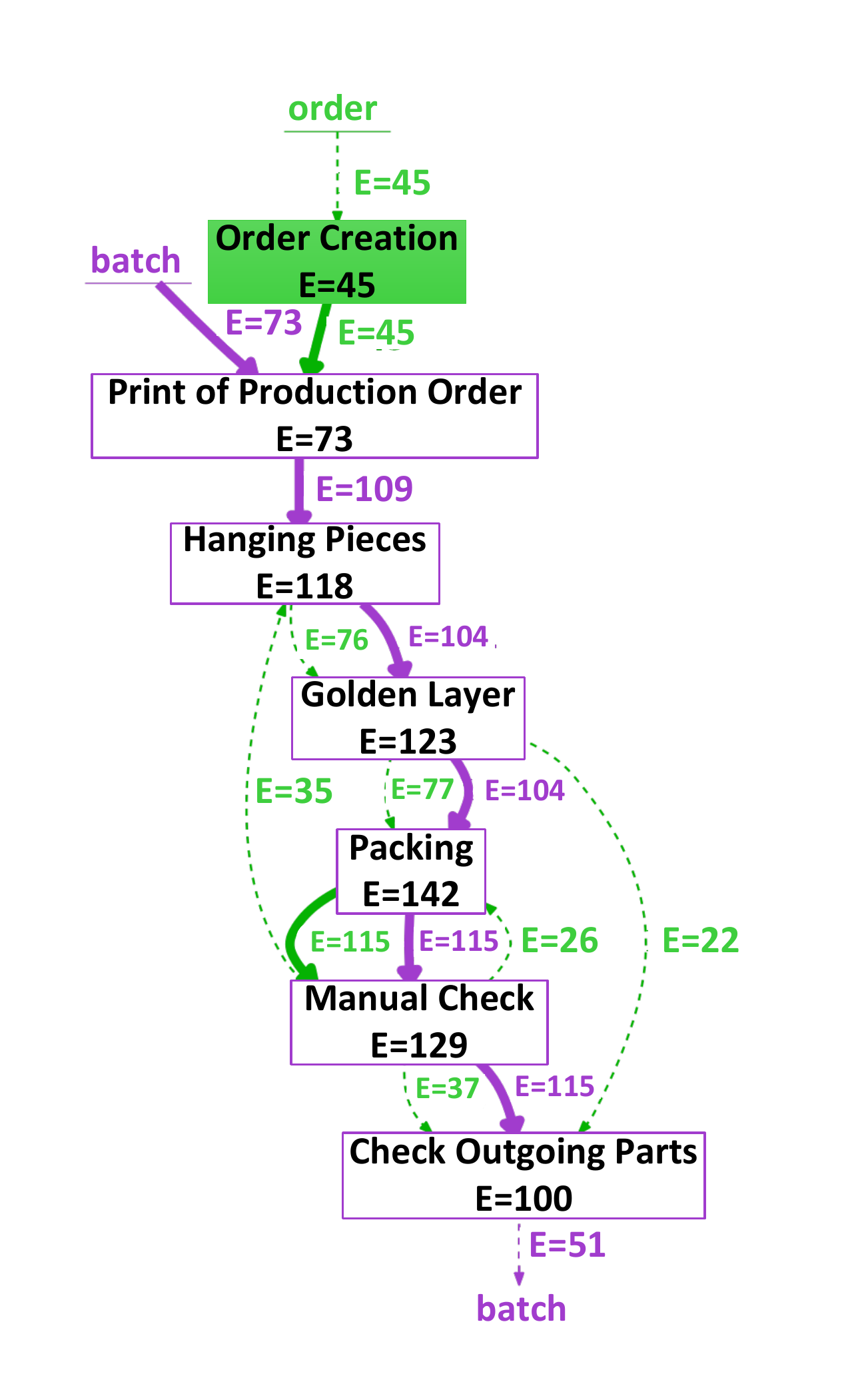}
    \caption{Cluster~3.}
  \end{subfigure}
  \begin{subfigure}[b]{0.57\linewidth}
    \includegraphics[width=\linewidth]{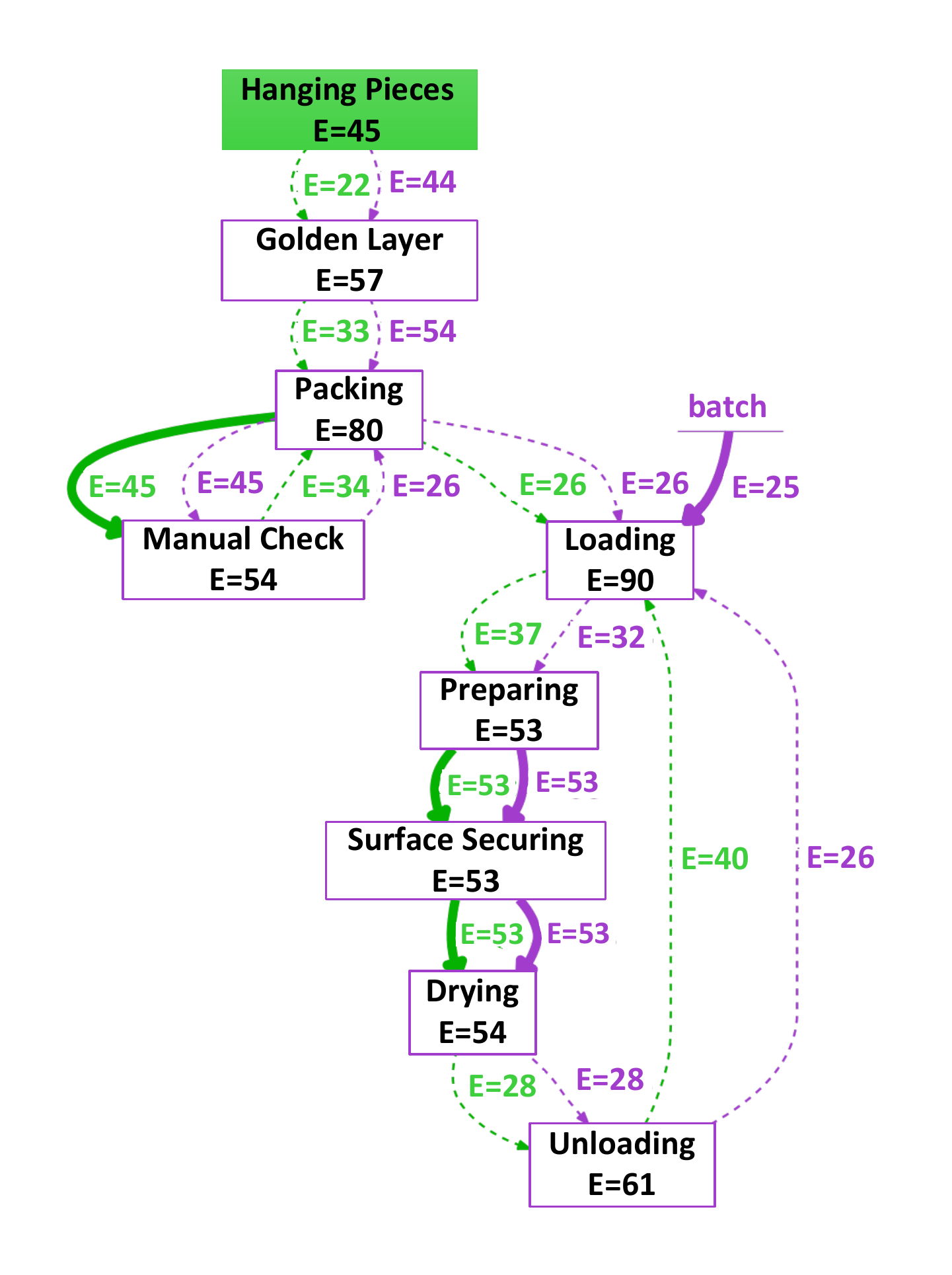}
    \caption{Cluster~4.}
  \end{subfigure}
  \caption{Final result of the clustering of batch based on \emph{all} approach.}
  \label{batch4}
\end{figure}

\section{Conclusion}\label{conclusion}

Process mining techniques provide valuable insights about process executions, however, most of the process mining techniques developed until now, focus on traditional event logs (i.e, event logs with single case notion). In reality, there exist processes with multiple interacting objects which are investigated in a new branch of process mining called object-centric process mining. Several process discovery techniques such as artifact modeling and Object-Centric Directly Follows Graphs (OC-DFGs) discovery have been developed to discover process models from object-centric processes, but the discovered models usually suffer from complexity. Therefore, in this paper, we propose an approach to obtain meaningful process models by clustering objects in an object-centric event log (OCEL). We introduce two approaches (i.e., \emph{all} and \emph{existence}) and use them in log extraction from the clusters. Furthermore, we enriched the OCEL with some graph-related features such as centrality to enhance clustering results. Moreover, to measure the quality of the process models, we have introduced complexity measures to evaluate the quality of OC-DFG models. We have applied our approach on a real-life B2B log of a manufacturing company applying surface treatment operations (e.g., lubricating and polishing) on various items. The results are promising where discovered process models can distinguish the process of different item types. For future work, we aim to evaluate the proposed approach on additional real data sets and use various quality metrics to evaluate the quality of the obtained process models more precisely\footnote{\scriptsize\textbf{Acknowledgments}: We thank the Alexander von Humboldt (AvH) Stiftung for supporting our research. Funded by the Deutsche Forschungsgemeinschaft (DFG, German Research Foundation) under Germany's Excellence Strategy–EXC-2023 Internet of Production – 390621612.}.

\bibliographystyle{plain}

\end{document}